\documentclass{article}
\usepackage{lmodern}
\usepackage{iclr2027_conference,times}
\usepackage{amsmath,amssymb,amsfonts,booktabs,graphicx}
\usepackage{float}
\usepackage{hyperref,url,xurl}
\usepackage[T1]{fontenc}
\hypersetup{hidelinks,pdfauthor={Xinkai Lu, Junli Zha, Haolan Wei, Jiahui Wang, Chao Zhou, Xiang Song},pdftitle={Redirecting Teacher Supervision: Temporal-Response-Guided Latent Distillation for Video-Language Models}}
\title{Redirecting Teacher Supervision: Temporal-Response-Guided Latent Distillation for Video-Language Models}
\author{Xinkai Lu \quad Junli Zha \quad Haolan Wei \quad Jiahui Wang \quad Chao Zhou \quad Xiang Song\\[3pt]
\normalfont SF Technology}
\iclrfinalcopy

\begin{document}

\maketitle
\fancyhead{}
\renewcommand{\headrulewidth}{0pt}

\begin{abstract}

On-policy distillation (OPD) transfers capabilities through teacher predictions on student-generated sequences, as studied by GKD and MiniLLM for language models and Video-OPD for video tasks. We ask how internal supervision can further support temporal learning. Direct hidden-state alignment does not explicitly use the student's temporal responses to choose a supervision direction. We propose Temporal-Response Guidance (TRG), which measures representation responses to local temporal perturbations, adds the teacher correction's projection onto their span back to the original correction, and restores its original norm. The projection is not normalized separately, and correction components outside the response space are retained. With an orthogonal tangent projector and a detached target, TRG changes the supervision direction while preserving the cosine-loss value and aligned-representation gradient norm at target construction. On TempCompass, Video-MME, MVBench, and Video-MMMU, TRG improves the 4B student's macro accuracy by 2.86 percentage points over the reported output-only OPD baseline.

\end{abstract}

\section{Introduction}

Video understanding depends on evidence across frames: the order of events, the direction of motion, and changes in state \citep{li2024mvbench}. Picking up and putting down a cup may look similar in individual frames, yet describe opposite events. TempCompass tests this distinction with videos that share static content but differ in temporal attributes \citep{liu2024tempcompass}. When transferring video understanding from a large teacher to a smaller model for deployment, we want the student to learn these temporal relations as well as recognize what appears in the scene.

On-policy distillation (OPD) offers a way to train on the answers a student actually generates. At each prefix of a student answer, the teacher provides a next-token distribution for supervision. GKD and MiniLLM study this approach in language models \citep{agarwal2024gkd,gu2024minillm}, and Video-OPD applies it to temporal video grounding \citep{li2026videoopd}. We build on this output objective and examine how representation supervision can help transfer temporal information.

Vanilla OPD compares token probabilities at the language modeling head \citep{agarwal2024gkd,gu2024minillm}. The hidden states before that head combine video evidence with the question and answer context. Although the output loss trains these states indirectly, it gives them no explicit representation target. Intermediate-feature distillation supplies such targets, with choices of feature transformation, layer, and distance metric affecting knowledge transfer \citep{heo2019overhaul}. Adding this supervision to OPD lets the student learn from the teacher's internal representations as well as its predictions.

Direct representation alignment uses the teacher state as its target. It does not use the student's response to temporal changes to decide which direction to emphasize. Input derivatives offer additional information about this response, as explored in Sobolev training and Jacobian matching \citep{czarnecki2017sobolev,srinivas2018jacobian}. For video distillation, we ask whether responses to controlled temporal changes can help choose the direction of the teacher correction. Doing so raises two questions:

\begin{itemize}

\item \textbf{Measuring temporal responses:} how can we isolate the student's response to local temporal changes from changes in the language context?

\item \textbf{Using partial response information:} how can we emphasize measured response directions while retaining correction components outside their span?

\end{itemize}

We propose \textbf{Temporal-Response Guidance (TRG)}. In its latent branch, both models receive the same teacher answer, and supervision is taken at a fixed position. We then perturb the video embeddings along structured temporal directions and use Jacobian-vector products (JVPs) to measure the student's first-order representation responses. Their span describes sensitivity to the selected perturbations under this fixed language context. The output loss is still computed on the student's own answer trajectories.

TRG projects the teacher correction onto this response space, adds the projection back to the original correction, and restores the original tangent norm. The size of the adjustment depends on how much the correction overlaps with the response space; components outside that space remain in the target. Under orthogonal tangent projection and stop-gradient, this changes the correction direction while matching direct alignment's loss and aligned-representation gradient norm at target construction (Proposition 1). Training gradients pass through the unperturbed student states. The response calculation is removed at inference.

\begin{figure}[!ht]

\centering

\includegraphics[width=\linewidth]{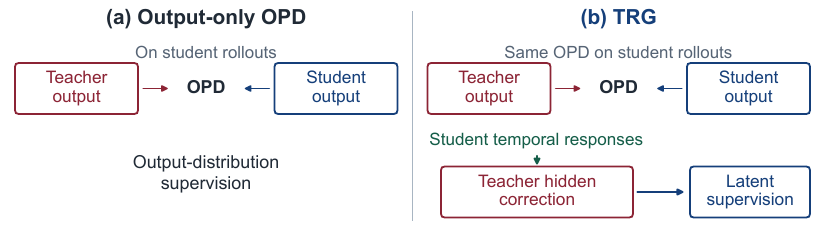}

\caption{Output-only OPD and TRG. Both use output supervision on student rollouts. TRG additionally uses student temporal responses to guide the teacher correction in latent space.}\label{fig:trg1}

\end{figure}

We evaluate a 27B teacher with 4B and 9B students on TempCompass, Video-MME, MVBench, and Video-MMMU \citep{liu2024tempcompass,fu2025videomme,li2024mvbench,hu2026videommmu}. At 4B, TRG improves macro accuracy by 2.86 points over OPD, and we also observe gains at 9B. Comparisons with direct alignment (Plain-AS) at both sizes and random correction orientations at 4B suggest a benefit from using student responses to construct the target. The paper makes three contributions:

\begin{itemize}

\item We introduce student responses to controlled local temporal changes as information for constructing video-distillation supervision.

\item We develop a residual projection rule that guides the teacher correction while retaining complementary components, and establish its local loss and representation-gradient scale properties.

\item We evaluate the design at two student scales on four benchmarks, using direct alignment, random orientation, and fixed-state geometry to examine response-guided supervision.

\end{itemize}

\section{Related Work}

\textbf{Output supervision and on-policy distillation.} Distillation methods choose both a target and the context in which the student learns from it. Sequence-level distillation trains on teacher-generated sequences \citep{kim2016sequence}; in classification, DKD separates target-class and non-target-class knowledge \citep{zhao2022dkd}. For autoregressive models, GKD trains on student-generated trajectories \citep{agarwal2024gkd}, while MiniLLM uses reverse KL to account for the student's limited capacity \citep{gu2024minillm}. Video-OPD extends dense on-policy supervision to temporal video grounding \citep{li2026videoopd}. Our work adds representation targets to this output objective.

\textbf{Latent supervision and target construction.} FitNets and attention transfer supervise intermediate features and attention maps \citep{romero2015fitnets,zagoruyko2017attention}. In language models, PKD transfers representations from multiple teacher layers \citep{sun2019patient}, and TinyBERT applies distillation during general pretraining and task learning \citep{jiao2020tinybert}. Feature transformations and cross-layer aggregation address the correspondence between teacher and student representations \citep{heo2019overhaul,chen2021review}. Other methods distill self-attention, inter-example distances and angles, or pairwise similarities \citep{wang2020minilm,park2019rkd,tung2019similarity}, and use variational or contrastive objectives to transfer shared information \citep{ahn2019vid,tian2020crd}. We fix the language context and layer correspondence, then use student temporal responses to choose how to reweight the teacher correction.

\textbf{Temporal structure in video learning.} Segment sampling and temporal shifts incorporate information from different times \citep{wang2016tsn,lin2019tsm}. Spatiotemporal attention models interactions across frames, while masked video modeling learns video representations through pretraining \citep{bertasius2021timesformer,arnab2021vivit,tong2022videomae}. TRG uses local changes in the video embeddings to measure student sensitivity, which then guides the representation target used for distillation.

\textbf{Local responses and derivative supervision.} Sobolev training and Jacobian matching use input derivatives in learning and knowledge transfer \citep{czarnecki2017sobolev,srinivas2018jacobian}. TRG computes student JVPs to identify a response subspace; it does not match student and teacher Jacobians. The teacher supplies the correction, and the student responses determine which of its components receives greater weight during training.

\section{Temporal-Response Guidance}

TRG has two supervision branches (Figure 2). OPD uses the student's generated answers for every training example. The latent branch uses accepted teacher answers on eligible videos, measures student temporal responses under that fixed context, and uses their span to redirect the teacher correction.

\begin{figure}[!ht]

\centering

\includegraphics[width=\linewidth]{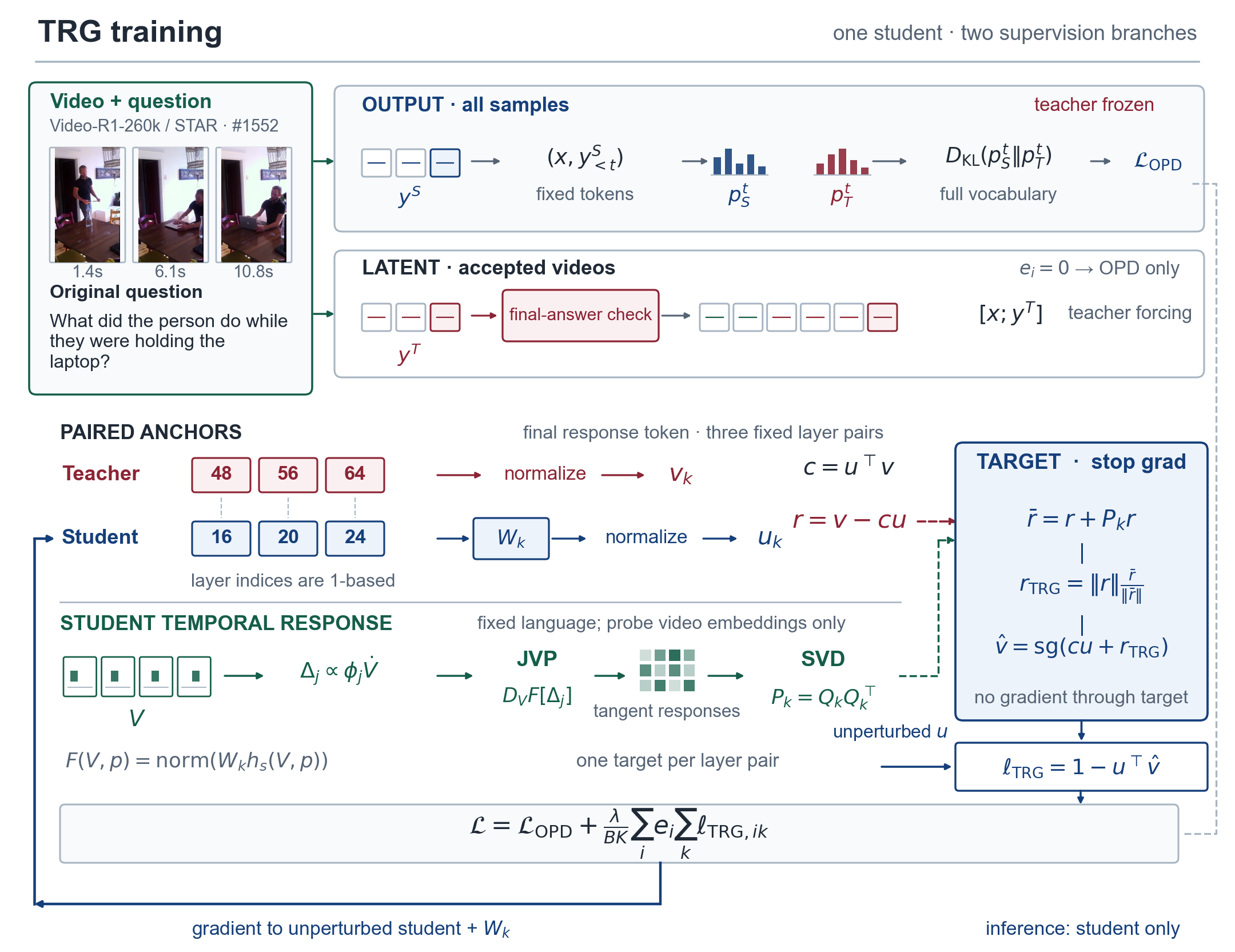}

\caption{TRG training framework. Frames and question: Video-R1-260k, STAR sample \#1552 \citep{feng2025videor1}. Student rollouts receive output supervision; accepted teacher answers fix the latent context. Responses construct a detached target, while gradients flow through unperturbed student states and alignment maps. Only the trained student is deployed.}\label{fig:trg2}

\end{figure}

\subsection{Supervision branches and the teacher correction}

\textbf{Output supervision.} Following OPD \citep{agarwal2024gkd}, let \(x\) contain an image or video and its question, and let \(y^S\) be the current student's generated answer. At answer position \(t\), both models receive \((x,y^S_{<t})\) and produce full-vocabulary next-token distributions \(p_T^t\) and \(p_S^t\). For batch size \(B\), answer length \(L_i\), and vocabulary \(\mathcal V\), we average reverse Kullback-Leibler (KL) divergence within each answer and then across examples:
\[\mathcal L_{\mathrm{OPD}}=\frac{1}{B}\sum_{i=1}^{B}\frac{1}{L_i}\sum_{t=1}^{L_i}\sum_{w\in\mathcal V}p_{S,i}^{t}(w)\log\frac{p_{S,i}^{t}(w)}{p_{T,i}^{t}(w)}.\tag{OPD}\]

We compute the loss on generated answer tokens, including end-of-sequence tokens and any tokens produced before truncation. Prompts and padding are excluded. The sampled token IDs and teacher predictions are fixed during differentiation, so gradients pass through the student predictions.

\textbf{Latent supervision.} The teacher generates a separate answer \(y^T\). A video contributes to the latent loss if this response is complete, untruncated, and its final answer matches the training label. We feed \([x;y^T]\) to both models using teacher forcing, so response measurements use the same language context. These videos receive both losses; images and the remaining videos receive only OPD (Appendix B).

The final teacher-answer token anchors supervision: it can attend to preceding video, question, and answer tokens. For each fixed layer pair \(\{(\ell_{s,k},\ell_{t,k})\}_{k=1}^{K}\), a trainable map \(W_k\in\mathbb R^{d\times d_s}\) aligns student state \(h_s^{(\ell_s)}\in\mathbb R^{d_s}\) to teacher state \(h_t^{(\ell_t)}\in\mathbb R^d\). Write \(z_s=W_kh_s^{(\ell_s)}\) and \(z_t=h_t^{(\ell_t)}\), omitting \(k\). Plain-AS and TRG share answers, anchors, and layer pairs. For nonzero \(z_s\) and \(z_t\), direct alignment minimizes
\[\ell(z_s,z_t)=1-\left(\frac{z_s}{\|z_s\|_2}\right)^\top\frac{z_t}{\|z_t\|_2}.\]

\textbf{Teacher correction.} Decompose the normalized teacher state into radial and tangent components along and orthogonal to the student direction, respectively. With identity matrix \(I_d\in\mathbb R^{d\times d}\),
\[\begin{aligned}c&=\frac{z_s^\top z_t}{\|z_s\|_2\|z_t\|_2},\\ r&=\left(I_d-\frac{z_s z_s^\top}{\|z_s\|_2^2}\right)\frac{z_t}{\|z_t\|_2}=\frac{z_t}{\|z_t\|_2}-c\frac{z_s}{\|z_s\|_2},\qquad a=\|r\|_2=\sqrt{1-c^2}.\end{aligned}\tag{1}\]

Here \(c\) is cosine similarity and \(a\) the tangent norm. The direct-alignment gradient is \(-r/\|z_s\|_2\), so descent follows \(r\), the \emph{teacher correction}. TRG redirects it using temporal responses while preserving its norm.

\subsection{Measuring student temporal responses}

Let \(V\in\mathbb R^{T\times N\times d_v}\) be video embeddings after visual encoding and projection, before entering the decoder: \(T\) temporal groups, \(N\) merged spatial tokens per group, and width \(d_v\). Groups need not correspond one-to-one with frames. Prompt and teacher-answer context \(p\), text embeddings, attention masks, token types, position IDs, and anchors remain fixed.

\textbf{Temporal probes.} A probe \(\Delta_j\in\mathbb R^{T\times N\times d_v}\) specifies an input direction. We estimate \(\dot V_t\) at corresponding spatial-grid positions using centered temporal differences internally and one-sided differences at boundaries. A time displacement \(\alpha w_t\) induces first-order change \(\alpha w_t\dot V_t\). We choose nonconstant discrete cosine transform (DCT) profiles for \(w\) and scale the resulting probes:
\[\begin{aligned}\phi_j(t)&=\sqrt{2/T}\cos\!\left(\pi(t+1/2)j/T\right),\\ \widetilde\Delta_{j,t}&=\phi_j(t)\dot V_t,\qquad\Delta_j=\frac{\|\dot V\|_F}{\|\widetilde\Delta_j\|_F+\epsilon}\widetilde\Delta_j.\end{aligned}\tag{2}\]

Here \(t=0,\ldots,T-1\), \(j=1,\ldots,m\), and \(m=T-1\). The Frobenius norm covers the full tensor; \(\epsilon>0\) stabilizes the denominator, and \(\|\dot V\|_F\) supplies a common scale. The nonconstant profiles span \(\{w\in\mathbb R^T:\mathbf1^\top w=0\}\). Omitting the constant (DC) profile excludes a uniform first-order shift and focuses on relative temporal deformations. These \(T-1\) profiles cover the chosen coefficient space. After modulation by \(\dot V_t\), the embedding probes need not be orthogonal or temporally centered. Appendix A discusses complete-basis equivalence and numerical truncation.

\textbf{Representation responses.} Define the normalized aligned state \(F(V,p)=z_s(V,p)/\|z_s(V,p)\|_2\in\mathbb R^d\). Its Jacobian \(D_VF(V,p)\) maps a video-embedding direction to a first-order representation change. We compute each JVP and collect tangent responses as
\[\begin{aligned}D_VF(V,p)[\Delta_j]&=\left.\frac{\mathrm d}{\mathrm d\alpha}F(V+\alpha\Delta_j,p)\right|_{\alpha=0},\\ d_j&=\left(I_d-\frac{z_s z_s^\top}{\|z_s\|_2^2}\right)D_VF(V,p)[\Delta_j],\qquad D=[d_1,\ldots,d_m]\in\mathbb R^{d\times m}.\end{aligned}\tag{2a}\]

Automatic differentiation evaluates JVPs with probes, model parameters, alignment maps, and context fixed at their current values. Since \(F\) is normalized, its exact derivative is tangent. Explicit projection removes radial numerical error, placing responses in the same tangent space as \(r\).

\textbf{Response subspace.} Singular value decomposition (SVD) of \(D\) supplies an orthonormal basis \(Q\), retaining left singular vectors with singular values at least \(\tau_{\mathrm{svd}}\) times the largest, \(0<\tau_{\mathrm{svd}}\le1\). Retained directions are not weighted; zero responses give an empty basis. The projector \(P=QQ^\top\) satisfies \(P=P^\top=P^2\) and \(Pz_s=0\) in exact arithmetic. Each layer pair uses its own basis and computes \(Pr=Q(Q^\top r)\) without a dense projector. Probes, responses, and bases are detached.

\subsection{Constructing the response-guided target}

TRG adds the response-space projection \(Pr\) back to the correction and restores its norm. If \(a=0\), we retain the teacher target; otherwise,
\[\bar r=r+Pr=2Pr+(I_d-P)r,\qquad r_{\mathrm{TRG}}=a\frac{\bar r}{\|\bar r\|_2}.\tag{3}\]

Equation (3) assigns weights of 2:1 to the response-space and complementary components before norm restoration. Because we use the projection at its original magnitude, little overlap with the response space produces only a small adjustment. The restored correction has norm \(\|r_{\mathrm{TRG}}\|_2=a\). A nonzero complementary component remains nonzero, though its magnitude changes. When the response space is empty, \(P=0\) and the target reduces to direct alignment.

Combining the redirected tangent correction with the original radial component gives the target and loss:
\[\widetilde z_t=c\frac{z_s}{\|z_s\|_2}+r_{\mathrm{TRG}},\qquad \ell_{\mathrm{TRG}}(z_s)=1-\frac{z_s^\top\operatorname{sg}(\widetilde z_t)}{\|z_s\|_2}.\tag{4}\]

The target has unit norm because the radial and tangent components are orthogonal and \(c^2+a^2=1\). Applying \(\operatorname{sg}\) detaches the target, while the unperturbed \(z_s\) remains differentiable. We construct a new target at the next update.

Let \(e_i\in\{0,1\}\) indicate an accepted video teacher answer, let \(\ell_{\mathrm{TRG},ik}\) be its loss at layer pair \(k\), and let \(\lambda\ge0\) weight latent supervision. The combined objective is
\[\mathcal L=\mathcal L_{\mathrm{OPD}}+\frac{\lambda}{BK}\sum_{i=1}^{B}e_i\sum_{k=1}^{K}\ell_{\mathrm{TRG},ik}.\tag{5}\]

Examples with \(e_i=0\) contribute only OPD; normalization still uses \(B\). Plain-AS uses target \(z_t/\|z_t\|_2\). Small corrections or projections trigger direct alignment (Appendix B). Training updates the student and alignment maps with a frozen teacher; inference retains only the student.

\subsection{Local geometric properties}

Consider a fixed construction point \(z_{s,0}\ne0\), a nonzero teacher state, an orthogonal tangent projector \(P\), and a detached target. All quantities below are evaluated at this point. Appendix A provides proofs; Appendix B describes finite-precision handling.

\textbf{Proposition 1 (loss and representation-gradient norm preservation).} TRG and direct cosine alignment have the same loss and gradient norm with respect to the aligned student state:
\[\begin{aligned}\ell_{\mathrm{TRG}}(z_{s,0})&=\ell(z_{s,0},z_t)=1-c,\\ \|\nabla_{z_s}\ell_{\mathrm{TRG}}(z_{s,0})\|_2&=\|\nabla_{z_s}\ell(z_{s,0},z_t)\|_2=\frac{a}{\|z_{s,0}\|_2}.\end{aligned}\tag{6}\]

The radial component preserves the inner product \(c\); tangent norm restoration preserves \(a/\|z_{s,0}\|_2\) while redirecting the gradient from \(-r\) to \(-r_{\mathrm{TRG}}\). Equal representation-gradient norms need not imply equal parameter-gradient norms.

\textbf{Proposition 2 (response-energy increase and bounded turning).} For \(a>0\), let \(q=\|Pr\|_2^2/a^2\) be the original response-space energy fraction and \(\theta\) the angle between \(r\) and \(r_{\mathrm{TRG}}\). Then
\[\frac{\|Pr_{\mathrm{TRG}}\|_2^2}{\|r_{\mathrm{TRG}}\|_2^2}=\frac{4q}{1+3q}\ge q,\qquad \cos\theta=\frac{1+q}{\sqrt{1+3q}},\quad 0\le\theta\le\arcsin(1/3).\tag{7}\]

Response energy increases for \(0<q<1\), retaining complementary components. The maximum turn is \(19.47^\circ\) at \(q=1/3\). Both endpoints leave direction unchanged: no component is emphasized at \(q=0\), and the whole correction is rescaled at \(q=1\). This bound follows from the residual weight, not an additional training threshold. Section 4 tests direction selection against Plain-AS and random orientation.

\section{Experiments}

We evaluate TRG against output-only OPD, then use direct alignment and random correction orientations to examine what response guidance adds. The TempCompass analysis locates the gains by question type and compares temporal probes with random input probes. Finally, we inspect how the target construction changes the teacher correction at fixed student states.

\subsection{Experimental setup}

\textbf{Models and training.} We follow Video-R1's data pipeline, first applying 3,000 SFT updates to Qwen3.5-4B and 9B students on Video-R1-CoT-165k, then distilling on Video-R1-260k, the dataset used for Video-R1's RL stage \citep{feng2025videor1}. The teacher is a previously fine-tuned Qwen3.5-27B video-CoT model and remains frozen. Distillation uses 300 AdamW updates, batch size 8, and learning rate \(10^{-4}\) to train attention LoRA parameters and, for latent supervision, the alignment maps \citep{loshchilov2019adamw,hu2022lora}. Plain-AS and TRG deploy the same student and adapter configuration. Appendix B gives the remaining training and evaluation settings.

\textbf{Data and repeated runs.} We use a deduplicated subset of 2,400 examples from Video-R1-260k, comprising 1,302 images and 1,098 videos. Every example contributes to OPD; videos with accepted teacher answers also contribute to the latent loss. The 300 updates make one pass through this fixed subset. The main comparisons at both student sizes and the random-orientation control at 4B each have three runs. These runs fix the SFT initialization, data subset, and sample order, varying rollout randomness. Appendix B specifies the input-length filter. Appendix D documents five source videos shared by the training data and MVBench and examines their scoring impact.

\textbf{Representations and probes.} Sixteen input frames produce \(T=8\) temporal groups and \(m=T-1=7\) nonconstant probes. Both student sizes use the relative SVD threshold \(\tau_{\mathrm{svd}}=10^{-3}\) and latent weight \(\lambda=1\). We align \(K=3\) layer pairs, \((16,48)\), \((20,56)\), and \((24,64)\), using one-based student and teacher indices.

\textbf{Benchmarks.} TempCompass tests temporal attributes such as order, direction, and speed \citep{liu2024tempcompass}. MVBench covers 20 video tasks involving temporal perception and reasoning \citep{li2024mvbench}. Video-MME spans a range of video domains and durations \citep{fu2025videomme}, while Video-MMMU tests perception, comprehension, and knowledge adaptation in multidisciplinary instructional videos \citep{hu2026videommmu}. We evaluate all four benchmarks with 16 frames; the sampling and answer-parsing rules are given in Appendix B.

\textbf{Baselines and reporting.} Vanilla OPD, abbreviated as \emph{OPD} in the tables, uses the output loss alone. Plain-AS adds direct cosine alignment with teacher states. It shares TRG's training examples, accepted teacher answers, supervision positions, layer pairs, and base objective; the two methods differ in how they construct the latent target. Tables 1 and 2 report mean accuracy over three runs. Macro accuracy gives each benchmark equal weight, and all means and differences are calculated from unrounded counts. Table C2 provides the counts for individual runs. The category and input-probe analysis in Table 3 uses one evaluation per method.

\subsection{Main results}

Table 1 compares the three distillation methods under our shared training and evaluation setup. Published video-model scores from Video-R1 are included for context \citep{feng2025videor1}; they use their original evaluation protocols.

\begin{table}[tbp]
\centering
\caption{Video-model accuracy (\%). Dagger-marked results come from Video-R1, Table 1, under their original protocols. Our 16-frame results are three-run means. Video-MME excludes subtitles. Dashes denote unlisted entries; macro scores cover only our controlled comparisons. TC: TempCompass; V-MME: Video-MME; MVB: MVBench; V-MMMU: Video-MMMU.}
\label{tab:1}
\small
\setlength{\tabcolsep}{4pt}
\begin{tabular}{lrrrrrr}
\toprule
Model / method & Frames & TC & V-MME & MVB & V-MMMU & Macro \\
\midrule
LongVA-7B\textsuperscript{\dag} & --- & 56.9 & 52.6 & --- & 23.9 & --- \\
VILA-1.5-8B\textsuperscript{\dag} & --- & 58.8 & --- & --- & 20.8 & --- \\
VILA-1.5-40B\textsuperscript{\dag} & --- & --- & 60.1 & --- & 34.0 & --- \\
Video-UTR-7B\textsuperscript{\dag} & --- & 59.7 & 52.6 & 58.8 & --- & --- \\
LLaVA-OneVision-7B\textsuperscript{\dag} & --- & --- & 58.2 & 56.7 & 33.8 & --- \\
Kangaroo-8B\textsuperscript{\dag} & --- & 62.5 & 56.0 & 61.1 & --- & --- \\
Video-R1-7B\textsuperscript{\dag} & 16 & 72.6 & 57.4 & 62.7 & 49.8 & --- \\
Video-R1-7B\textsuperscript{\dag} & 32 & 73.2 & 59.3 & 63.9 & 52.3 & --- \\
Video-R1-7B\textsuperscript{\dag} & 64 & 73.2 & 61.4 & 64.8 & 52.4 & --- \\
\midrule
4B OPD & 16 & 73.23 & 56.06 & 64.77 & 59.89 & 63.49 \\
4B Plain-AS & 16 & 74.24 & 58.70 & 64.33 & 64.67 & 65.48 \\
4B TRG & 16 & \textbf{74.81} & \textbf{60.67} & \textbf{65.03} & \textbf{64.89} & \textbf{66.35} \\
\midrule
9B OPD & 16 & 72.22 & 59.19 & 64.73 & 61.00 & 64.28 \\
9B Plain-AS & 16 & 74.87 & 61.72 & 64.68 & 65.33 & 66.65 \\
9B TRG & 16 & \textbf{75.38} & \textbf{63.52} & \textbf{65.30} & \textbf{65.56} & \textbf{67.44} \\
\bottomrule
\end{tabular}
\end{table}

TRG improves over OPD at both student sizes and on all four benchmarks (Table 1). Macro accuracy rises by 2.86 points at 4B and 3.16 points at 9B. The improvement is largest on Video-MME and Video-MMMU, with a smaller gain on MVBench. The benefit of the latent branch is therefore visible at both model sizes, although its magnitude depends on the benchmark.

\subsection{Contribution of response guidance}

\textbf{Direct representation alignment.} Plain-AS already improves macro accuracy over OPD at both student sizes, showing that representation supervision contributes to the overall gain. TRG adds a further 0.87 points at 4B and 0.79 at 9B. Since Plain-AS and TRG share the teacher answers, supervision positions, layer pairs, and base objective, this comparison supports using student responses to construct the latent target. The gain is positive in each corresponding run (Table C2).

The additional benefit from TRG is largest on Video-MME. On Video-MMMU, most of the improvement over OPD is already obtained by direct alignment. MVBench shows a different pattern: Plain-AS falls slightly below OPD, while TRG exceeds both. Thus, adding a latent loss helps the macro average, but its benefit on individual benchmarks depends on the target.

\textbf{Correction direction.} The random-orientation control tests whether the direction selected from student responses matters. It applies TRG's angle rule at its own current anchor, then chooses a random tangent orientation. For a nondegenerate anchor with \(a>0\), set \(g=r/a\) and \(d_{\mathrm{rand}}=\cos\theta\,g+\sin\theta\,\xi\), where \(\xi\) is a random unit vector orthogonal to both \(z_s/\|z_s\|_2\) and \(g\). The target \(c z_s/\|z_s\|_2+ad_{\mathrm{rand}}\) preserves the ideal local loss and representation-gradient norm. Responses determine the turn angle through the shared rule, but no longer determine its orientation.

\begin{table}[tbp]
\centering
\caption{Angle-rule-matched direction comparison at 4B. Accuracy (\%), averaged over three runs. Shared rows reuse Table 1.}
\label{tab:2}
\small
\setlength{\tabcolsep}{4pt}
\begin{tabular}{lrrrrr}
\toprule
Method & TC & V-MME & MVB & V-MMMU & Macro \\
\midrule
Plain-AS & 74.24 & 58.70 & 64.33 & 64.67 & 65.48 \\
Random orientation & 74.43 & 59.19 & 64.80 & 63.22 & 65.41 \\
TRG & \textbf{74.81} & \textbf{60.67} & \textbf{65.03} & \textbf{64.89} & \textbf{66.35} \\
\bottomrule
\end{tabular}
\end{table}

TRG exceeds this control by 0.94 macro points at 4B and has higher mean accuracy on each benchmark (Table 2). The difference favors response-guided directions over random orientations under the same turning rule. This is a more specific comparison than TRG versus Plain-AS, where the correction is not redirected.

\subsection{Temporal categories and probe design}

TempCompass separates action recognition from direction, speed, event order, and attribute change, allowing us to examine which questions benefit from response guidance. Table 3 reports one evaluation per method at 4B. Alongside OPD and Plain-AS, it includes a random-input-response control that changes the probes used to measure student sensitivity.

\begin{table}[tbp]
\centering
\caption{TempCompass categories, 4B and 16 frames. Accuracy (\%) from one evaluation per method, separate from Table 1. Headers give question counts (1,580 total). The last row reports count-based percentage-point gains; complete counts appear in Appendix C.}
\label{tab:3}
\small
\setlength{\tabcolsep}{4pt}
\begin{tabular}{lrrrrrr}
\toprule
Method & \shortstack{Action\\(338)} & \shortstack{Direction\\(335)} & \shortstack{Speed\\(317)} & \shortstack{Order\\(302)} & \shortstack{Attr. change\\(288)} & Overall \\
\midrule
OPD & 80.47 & 71.34 & 72.56 & 68.87 & 72.22 & 73.23 \\
Plain-AS & \textbf{81.07} & 72.54 & 73.50 & 70.20 & 73.26 & 74.24 \\
Random-input response & 80.77 & 72.54 & 73.82 & 69.54 & 73.26 & 74.11 \\
TRG & \textbf{81.07} & \textbf{73.13} & \textbf{74.13} & \textbf{71.19} & \textbf{73.96} & \textbf{74.81} \\
\midrule
TRG - Plain-AS (pp) & 0.00 & +0.60 & +0.63 & +0.99 & +0.69 & +0.57 \\
\bottomrule
\end{tabular}
\end{table}

The clearest additional gain is on event-order questions. Relative to Plain-AS, TRG improves this category by 0.99 points, while action accuracy is unchanged; direction, speed, and attribute change also improve. The advantage over direct alignment is therefore concentrated in temporal relations and state changes in this evaluation. Event order also shows the largest gain over OPD (+2.32 points), yet remains the lowest-scoring category at 71.19\%.

\textbf{Probe design.} To examine the temporal probes, we replace them with Gaussian perturbations of the video embeddings. The random probes are centered over time and matched to the temporal probes' Frobenius norms. JVP measurement, subspace extraction, and target construction remain unchanged (Appendix C). This comparison changes the inputs used to obtain student responses; the direction control in Section 4.3 instead changes how those responses guide the correction.

TRG gains 0.70 points over the random-input-response control, with the largest difference again on event order (+1.66 points). This favors the chosen temporal probes in the present evaluation, but their advantage across runs has not been measured. Table C1 gives the remaining single-run comparisons of target construction and probe count.

\subsection{Correction geometry at fixed states}

We examine how much TRG changes the teacher correction using archived representations from 789 examples. Three layer pairs per example give 2,367 sample-layer anchors. At each anchor, we compare the original correction \(r\) with \(r_{\mathrm{TRG}}\) at the same student state. Response-space energy is the fraction of the correction's squared norm lying in the measured response space.

The correction turns by \(6.07^\circ\) on average. The norm of the change, measured relative to the original correction norm, averages 10.58\%. Meanwhile, the mean response-space energy increases from 1.49\% to 5.48\%. Each quantity is computed per anchor before averaging. On these states, TRG therefore places more weight on the measured response directions while making a relatively small change to the teacher correction.

This behavior follows from adding the projection in (3). The response-space component is doubled before norm restoration, while the complementary component is retained. Both are then rescaled by the same factor to restore the original norm. The low response-space energy after reweighting shows that much of the correction still lies outside the measured span. A projection-only target would remove that complementary component; TRG keeps it in the supervision target.

\section{Discussion and Limitations}

The gains over Plain-AS suggest that the benefit of latent supervision depends in part on how its target is constructed. TRG uses the student's current sensitivity to reweight the teacher correction, while retaining components outside the measured span. The benchmark differences show that the additional benefit is uneven: direct alignment accounts for most of the gain on Video-MMMU, whereas response guidance contributes more on Video-MME.

The random-orientation result at 4B favors using measured student responses to choose the correction direction. The random-input control asks a narrower question about the temporal probes. Its single TempCompass evaluation favors the chosen probes, but does not establish that this advantage persists across runs or on other benchmarks.

Our evaluation covers one teacher family and two student sizes. Repeated runs vary rollout randomness while keeping the SFT initialization, data subset, and sample order fixed. The choice of teacher-answer context has not been compared with other language contexts. The theoretical results concern local loss and representation-gradient norms at target construction; convergence as the targets and model parameters change remains outside this analysis.

TRG leaves the inference model unchanged, but JVPs and subspace construction add computation during training. The experiments compare methods at the same number of updates. Performance under equal compute budgets remains untested.

\section{Conclusion}

Our experiments suggest that student temporal responses can help choose the direction of representation supervision in video distillation. We proposed TRG, which measures these responses under a fixed language context and uses their span to reweight the teacher correction, retaining components outside the measured response space. Across four benchmarks, TRG improves on output-only OPD and direct representation alignment with both 4B and 9B students. At 4B, it also outperforms random correction orientations under the same angle rule. The response computation is used during training, leaving the model used at inference unchanged.

\label{main-text-end}

\clearpage

\section*{AI Use Statement}

Generative AI tools assisted with writing experimental code and gathering research materials. The authors take responsibility for the final text, claims, and results.

\bibliographystyle{iclr2027_conference}

\bibliography{references_submission}

\clearpage

\appendix

\section{Proofs and correction geometry}

\subsection{Proofs of Propositions 1 and 2}

At the construction point \(z_{s,0}\ne0\), write \(r_\parallel=Pr\) and \(r_\perp=(I_d-P)r\). Orthogonality gives \(\bar r=2r_\parallel+r_\perp\) and \(\|\bar r\|^2=4\|r_\parallel\|^2+\|r_\perp\|^2>0\) for \(r\ne0\). Norm restoration makes \(r_{\mathrm{TRG}}\) tangent with norm \(a\), so \(\widetilde z_t\) has unit norm and inner product \(c\) with \(z_{s,0}/\|z_{s,0}\|_2\).

For any fixed unit target \(b\), the gradient of \(1-z_s^\top b/\|z_s\|_2\) is \(-(I_d-z_sz_s^\top/\|z_s\|_2^2)b/\|z_s\|_2\). Substituting the direct and redirected targets gives \(-r/\|z_{s,0}\|_2\) and \(-r_{\mathrm{TRG}}/\|z_{s,0}\|_2\). Their norms and loss values agree, proving Proposition 1. If \(r=0\), direct-alignment fallback gives the same result.

For \(a>0\) and \(q=\|r_\parallel\|^2/a^2\), we have \(\|\bar r\|^2=a^2(1+3q)\), \(\|P\bar r\|^2=4a^2q\), and \(r^\top\bar r=a^2(1+q)\). These establish both expressions in (7). The energy-fraction increase is \(3q(1-q)/(1+3q)\). The derivative of \((1+q)/\sqrt{1+3q}\) is \((3q-1)/(2(1+3q)^{3/2})\). Its zero at \(q=1/3\) gives the minimum cosine \(2\sqrt2/3\), hence maximum turn \(\arcsin(1/3)\), proving Proposition 2.

\subsection{Probe space and basis choice}

Let \(\mathcal H=\{w\in\mathbb R^T:\mathbf1^\top w=0\}\) and \(\mathcal A_V(w)_t=w_t\dot V_t\). Composing \(\mathcal A_V\) with the tangent-projected student Jacobian gives a linear map \(\mathcal M:\mathcal H\to\mathbb R^d\). For DCT basis \(\Phi\), the unscaled response matrix is \(D_0=\mathcal M\Phi\). Any complete alternative basis is \(\Phi R\) for invertible \(R\), giving \(D'_0=D_0R\) and the same column space. Positive per-probe scaling also preserves this span; when \(\dot V=0\), all responses vanish. With relative SVD truncation, however, the retained subspace can depend on the basis.

The \(T-1\) probes span the chosen coefficient family; their coverage does not extend to arbitrary embedding perturbations. Removing the DC profile excludes a uniform first-order temporal shift. Because \(\dot V_t\) varies with time, zero-mean coefficients can still produce embedding probes with a nonzero temporal mean. The random-input control consequently tests a different probe construction, rather than a change of basis within the same subspace.

\subsection{Alternative target geometry}

For \(a>0\), let \(g=r/a\) and \(\rho=\|Pg\|\in(0,1]\). Then the projection-only and unit-projection correction directions make angles of \(\arccos\rho\) and \(\tfrac12\arccos\rho\) with \(g\), respectively. TRG satisfies
\[\cos^2\theta_{\mathrm{TRG}}-\frac{1+\rho}{2}=\frac{(1-\rho)^2(2\rho^2+\rho+1)}{2(1+3\rho^2)}\ge0.\tag{8}\]

Hence \(\theta_{\mathrm{TRG}}\le\theta_{\mathrm{unit}}\le\theta_{\mathrm{proj}}\). The TRG angle vanishes as \(\rho\to0\), whereas the alternatives approach \(45^\circ\) and \(90^\circ\).

On the anchors in Section 4.5, TRG's mean turn is \(6.07^\circ\). With the shared numerical cap of approximately \(21.99^\circ\) specified in Appendix B, the mean TRG angle is unchanged. The cap reduces the unit-projection mean from \(41.84^\circ\) to \(21.99^\circ\) and the projection-only mean from \(83.68^\circ\) to \(21.99^\circ\). The two alternatives produce the same clipped directions on these anchors.

\section{Implementation and evaluation}

\textbf{Training.} Table B1 gives the shared distillation settings. Students receive 3,000 SFT updates; the previously fine-tuned video-CoT teacher stays frozen. Repeated runs use the same initialization, 2,400-example subset, and sample order. The input-length filter is applied after preprocessing.

\renewcommand{\thetable}{B1}

\begin{table}[!htb]

\centering

\caption{Distillation settings shared by 4B and 9B. Layer indices are one-based.}\label{tab:B1}

\small

\setlength{\tabcolsep}{3.5pt}

\begin{tabular}{@{}p{.50\linewidth}p{.46\linewidth}@{}}

\toprule

Setting & Value \\

\midrule

Video frames / temporal groups / probes & 16 / 8 / 7 \\

Updates / effective batch size & 300 / 8 \\

Learning rate / weight decay & \(10^{-4}\) / 0.01 \\

Attention LoRA rank / scale / dropout & 16 / 32 / 0 \\

Student-teacher layer pairs & (16,48), (20,56), (24,64) \\

Relative SVD threshold / latent weight & \(10^{-3}\) / 1 \\

Maximum input / output tokens & 2,048 / 1,024 \\

\bottomrule

\end{tabular}

\end{table}

Student rollout uses temperature 1, top-p 1, and no top-k truncation; the prompt requests answer tags. The teacher generates one greedy response per candidate. Plain-AS and TRG share complete teacher responses whose parsed answers match the training labels. The output loss averages response-token reverse KL within examples and then across the batch. The latent loss averages three layers per accepted video and uses the full-batch denominator, as in (5).

\begin{samepage}

\textbf{Numerics and cost.} JVPs use forward-mode automatic differentiation in FP32, with TF32 and dropout disabled. Responses are tangent-projected and detached before SVD. Direct alignment is used if rank is zero, \(a<10^{-4}\), or \(\|Pg\|<10^{-3}\). Cosines are clipped to \([-1+10^{-6},1-10^{-6}]\). The \(21.994192^\circ\) numerical angle cap exceeds the ideal TRG bound, but can clip alternative targets. Norm restoration precedes target detachment; optimization differentiates only through unperturbed states. For dimension \(d\), \(m\) probes, and retained rank \(k\), response storage, SVD, and projection cost \(O(dm)\), \(O(dm^2)\), and \(O(dk)\), excluding model passes and JVPs. A probe pass can collect several layers; SVD rank truncation does not reduce JVP cost.

\end{samepage}

\renewcommand{\thetable}{B2}

\begin{table}[!htb]

\centering

\caption{Evaluation uses 16 frames, greedy decoding, thinking disabled, and no audio or subtitles. Invalid or non-EOS-truncated answers are incorrect.}\label{tab:B2}

\small

\setlength{\tabcolsep}{3.5pt}

\begin{tabular}{@{}lp{.44\linewidth}rr@{}}

\toprule

Benchmark & Frame sampling & Output limit & Questions \\

\midrule

TempCompass & Processor fixed-frame, full video & 512 & 1,580 \\

Video-MME & Uniform, full video & 512 & 2,700 \\

MVBench & Official segment centers and clip bounds & 512 & 4,000 \\

Video-MMMU & Uniform; Adaptation question image & 4,096 & 900 \\

\bottomrule

\end{tabular}

\end{table}

TempCompass and Video-MME use exact option matching, accepting unambiguous answer tags, explicit final answers, initial option letters, or final-paragraph commitments. MVBench adds terminal-letter recovery and option-range checks; frame folders use official 3-FPS, one-based indexing. Video-MMMU uses author parsers (revision 88b4ac3e0732), including open-answer matching. Perception/Comprehension prompts ignore embedded quizzes; one untyped ten-option question is treated as multiple choice.

\section{Supplementary results and probe control}

\textbf{Single-run variants.} Each row in Table C1 reports one run. Its macro score averages the displayed accuracies; main-table macros use unrounded count-based scores. Unit projection adds normalized \(Pg\) to \(g\), whereas projection only uses \(Pg/\|Pg\|\). Both alternatives change the turn angle and component weights. Fewer probes require fewer JVPs, but these runs do not measure end-to-end speedup or variability across runs.

\renewcommand{\thetable}{C1}

\begin{table}[!htb]

\centering

\caption{Single-run variants at 4B; accuracy (\%).}\label{tab:C1}

\small

\setlength{\tabcolsep}{3.5pt}

\begin{tabular}{@{}lrrrrr@{}}

\toprule

Variant & TC & V-MME & MVB & V-MMMU & Macro \\

\midrule

Plain \(\lambda=0.3\) & 73.61 & 57.19 & 64.70 & 62.33 & 64.46 \\

Plain \(\lambda=3\) & 74.18 & 58.81 & 64.30 & 64.56 & 65.46 \\

Unit projection & 74.87 & 59.41 & 64.75 & 64.33 & 65.84 \\

Projection only & 73.92 & 58.52 & 64.05 & 63.67 & 65.04 \\

3 probes & 74.62 & 60.11 & 64.93 & 64.56 & 66.06 \\

5 probes & 74.75 & 60.52 & 65.00 & 64.78 & 66.26 \\

\bottomrule

\end{tabular}

\end{table}

\textbf{Repeated-run counts.} Table C2 gives the counts underlying Tables 1 and 2. Each accuracy is the correct count divided by the benchmark total. Macro scores average the four benchmark accuracies and then the three runs, without intermediate rounding.

\renewcommand{\thetable}{C2}

\begin{table}[!htb]

\centering

\caption{Correct counts in run order (1, 2, 3); headers give evaluation totals.}\label{tab:C2}

\footnotesize

\setlength{\tabcolsep}{3.5pt}

\begin{tabular}{@{}lrrrr@{}}

\toprule

Scale / method & TC /1580 & V-MME /2700 & MVB /4000 & V-MMMU /900 \\

\midrule

4B OPD & 1164, 1151, 1156 & 1532, 1484, 1525 & 2588, 2581, 2603 & 538, 545, 534 \\

4B Plain-AS & 1171, 1177, 1171 & 1588, 1584, 1583 & 2572, 2582, 2565 & 581, 583, 582 \\

4B Random orientation & 1180, 1181, 1167 & 1618, 1594, 1582 & 2593, 2594, 2589 & 570, 571, 566 \\

4B TRG & 1186, 1180, 1180 & 1641, 1639, 1634 & 2600, 2599, 2605 & 586, 583, 583 \\

\midrule

9B OPD & 1139, 1139, 1145 & 1599, 1595, 1600 & 2594, 2586, 2588 & 548, 548, 551 \\

9B Plain-AS & 1186, 1181, 1182 & 1672, 1663, 1664 & 2585, 2586, 2591 & 589, 586, 589 \\

9B TRG & 1193, 1193, 1187 & 1709, 1719, 1717 & 2611, 2616, 2609 & 591, 591, 588 \\

\bottomrule

\end{tabular}

\end{table}

\textbf{Random-input control.} Seven standard-normal probes have the same shape as the video embeddings. Each is centered over eight temporal groups and rescaled to its corresponding temporal probe's Frobenius norm. JVPs, SVD, numerical safeguards, and detached target construction are unchanged from TRG. Table C3 gives the category counts for Table 3, using one evaluation per method.

\renewcommand{\thetable}{C3}

\begin{table}[!htb]

\centering

\caption{Single-evaluation TempCompass correct counts underlying Table 3.}\label{tab:C3}

\footnotesize

\setlength{\tabcolsep}{3.5pt}

\begin{tabular}{@{}lrrrrrr@{}}

\toprule

Method & \shortstack{Action\\/338} & \shortstack{Direction\\/335} & \shortstack{Speed\\/317} & \shortstack{Order\\/302} & \shortstack{Change\\/288} & \shortstack{Total\\/1580} \\

\midrule

OPD & 272 & 239 & 230 & 208 & 208 & 1157 \\

Plain-AS & 274 & 243 & 233 & 212 & 211 & 1173 \\

Random-input response & 273 & 243 & 234 & 210 & 211 & 1171 \\

TRG & 274 & 245 & 235 & 215 & 213 & 1182 \\

\bottomrule

\end{tabular}

\end{table}

\section{Media overlap and score sensitivity}

File hashes identify five byte-identical videos shared by six LLaVA-Video/Charades training records and five STAR/Charades MVBench questions (Table D1). The questions differ between training and evaluation. We verified the shared source files, but not whether the crops or sampled frames coincide. Overlap among the remaining media has not been ruled out.

\renewcommand{\thetable}{D1}

\begin{table}[!htb]

\centering

\caption{Shared media and affected MVBench items.}\label{tab:D1}

\small

\setlength{\tabcolsep}{3.5pt}

\begin{tabular}{@{}lll@{}}

\toprule

Video & MVBench item & Training record IDs \\

\midrule

8LAK1.mp4 & action\_prediction:0040 & 184504, 50183 \\

KQYR0.mp4 & action\_prediction:0105 & 181881 \\

FQS7O.mp4 & action\_sequence:0017 & 27574 \\

ZFQNZ.mp4 & action\_sequence:0148 & 229025 \\

3EIRZ.mp4 & object\_interaction:0047 & 85670 \\

\bottomrule

\end{tabular}

\end{table}

In one audited TRG prediction file, four of the five affected items are correct. Removing all five changes accuracy from 2,601/4,000 (65.025\%) to 2,597/3,995 (65.006\%), a decrease of 0.019 points. Allowing arbitrary correctness on these items and rounding of the displayed means gives conservative TRG-minus-Plain-AS gaps above 0.56 points at 4B and 0.48 at 9B. These are deletion bounds, not rerun scores.

Excluding MVBench gives three-benchmark macro gains of 0.92 and 0.84 points (four-benchmark gains: 0.87 and 0.79). These checks bound direct scoring effects, not training-exposure effects; MVBench results remain specific to this data setting.

\end{document}